# *Attend to You*: Personalized Image Captioning with Context Sequence Memory Networks

Cesc Chunseong Park*
Lunit Inc.
Seoul, Korea
cspark@lunit.io

Byeongchang Kim    Gunhee Kim
Seoul National University
Seoul, Korea
byeongchang.kim@vision.snu.ac.kr    gunhee@snu.ac.kr

## Abstract

*We address personalization issues of image captioning, which have not been discussed yet in previous research. For a query image, we aim to generate a descriptive sentence, accounting for prior knowledge such as the user's active vocabularies in previous documents. As applications of personalized image captioning, we tackle two post automation tasks: hashtag prediction and post generation, on our newly collected Instagram dataset, consisting of 1.1M posts from 6.3K users. We propose a novel captioning model named Context Sequence Memory Network (CSMN). Its unique updates over previous memory network models include (i) exploiting memory as a repository for multiple types of context information, (ii) appending previously generated words into memory to capture long-term information without suffering from the vanishing gradient problem, and (iii) adopting CNN memory structure to jointly represent nearby ordered memory slots for better context understanding. With quantitative evaluation and user studies via Amazon Mechanical Turk, we show the effectiveness of the three novel features of CSMN and its performance enhancement for personalized image captioning over state-of-the-art captioning models.*

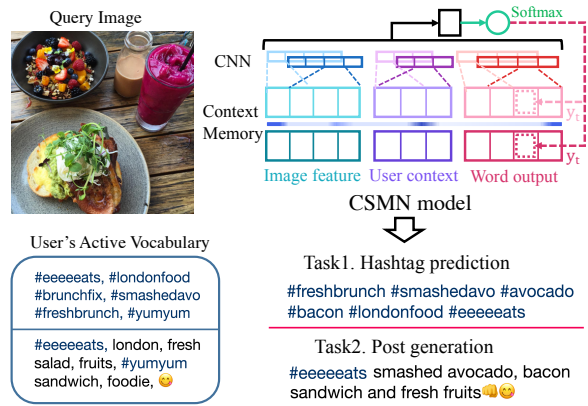

Figure 1. Problem statement of personalized image captioning with an Instagram example. As main applications, we address hashtag prediction and post generation tasks. Given a query image, the former predicts a list of hashtags, while the latter generates a descriptive text to complete a post. We propose a versatile *context sequence memory network* (*CSMN*) model.

## 1. Introduction

Image captioning is a task of automatically generating a descriptive sentence of an image [3, 4, 9, 12, 20, 22, 28, 31, 33]. As this task is often regarded as one of frontier-AI problems, it has been actively studied in recent vision and language research. It requires an algorithm not only to understand the image content in depth beyond category or attribute levels, but also to connect its interpretation with a language model to create a natural sentence.

This work addresses *personalization* issues of image captioning, which have not been discussed in previous research. We aim to generate a descriptive sentence for an image, accounting for prior knowledge such as the user's active vocabularies or writing styles in previous documents. Potentially, personalized image captioning is applicable to a wide range of automation services in photo-sharing social networks. For example, in Instagram or Facebook, users instantly take and share pictures as posts using mobile phones. One bottleneck to complete an image post is to craft hashtags or associated text description using their own words. Indeed, *crafting text* is more cumbersome than *taking a picture* for general users; photo-taking can be done with only a single tab on the screen of a smartphone, whereas text writing requires more time and mental energy for selecting suitable keywords and completing a sentence to describe theme, sentiment, and context of the image.

In this paper, as examples of personalized image captioning, we focus on two post automation tasks: *hashtag*

---

*prediction* and *post generation*. Figure 1 shows an Instagram post example. The hashtag prediction automatically predicts a list of hashtags for the image, while the post generation creates a sentence consisting of normal words, emojis, and even hashtags. Personalization is key to success in these two tasks, because text in social networks is not simple description of image content, but the user's own story and experience about the image with his or her favorite vocabularies and expressions.

To achieve personalized image captioning tasks, we propose a memory network model named as *context sequence memory network* (*CSMN*). Our model is inspired by *memory networks* [5, 24, 29], which explicitly include memory components to which neural networks read and write data for capturing long-term information. Our major updates over previous memory network models are three-fold.

First, we propose to use the memory as a context repository of prior knowledge for personalized image captioning. Since the topics of social network posts are too broad and users' writing styles are too diverse, it is crucial to leverage prior knowledge about the authors or metadata around the images. Our memory retains such multiple types of context information to promote more focused prediction, including users' active vocabularies and various image descriptors.

Second, we design the memory to sequentially store all of the words that the model generates. It leads two important advantages. First, it enables the model to selectively attend, at every step, on the most informative previous words and their combination with other context information in the memory. Second, our model does not suffer from the vanishing gradient problem. Most captioning models are equipped with RNN-based encoders (*e.g.* [3, 22, 25, 28, 31, 33]), which predict a word at every time step, based on only a current input and a single or a few hidden states as an implicit summary of all previous history. Thus, RNNs and their variants often fails to capture long-term dependencies, which could worsen if one wants to use prior knowledge together. On the other hand, our state-based sequence generation explicitly retains all information in the memory to predict next words. By using teacher-forced learning [30], our model has a markov property at training time; predicting a previous word $y_{t-1}$ has no effect on predicting a next word $y_t$, which depends on only the current memory state. Thus, the gradients from the current time step prediction $y_t$ are not propagated through the time.

Third, we propose to exploit a CNN to jointly represent nearby ordered memory slots for better context understanding. Original memory networks [24, 29] leverage time embedding to model the memory order. Still its representation power is low, since it cannot represent the correlations between multiple memory slots, for which we exploit convolution layers that lead much stronger representation power.

For evaluation, we collect a new personalized image captioning dataset, comprising 1.1M Instagram posts from 6.3K users. Instagram is a great source for personalized captioning, because posts mostly include personal pictures with long hashtag lists and characteristic text with a wide range of topics. For each picture post, we consider the body text or a list of hashtags as groundtruth captions.

Our experimental results demonstrate that aforementioned three unique features of our CSMN model indeed improve captioning performance, especially for personalization purpose. We also validate that our CSMN significantly outperforms several state-of-the-art captioning models with the decoders of RNNs or LSTMs (*e.g.* [27, 28, 31]). We evaluate with quantitative language metrics (*e.g.* BLEU [21], CIDEr [26], METEOR [14], and ROUGE [15]) and user studies via Amazon Mechanical Turk.

We summarize contribution of this work as follows.

(1) To the best of our knowledge, we propose a first *personalized* image captioning approach. We introduce two practical post automation tasks that benefit from personalized captioning: post generation and hashtag prediction.

(2) We propose a novel memory network model named CSMN for personalized captioning. The unique updates of CSMN include (i) exploiting memory as a repository for multiple context information, (ii) appending previously generated words into memory to capture long-term information without, and (iii) adopting CNN memory structure to jointly represent nearby ordered memory slots.

(3) For evaluation of personalized image captioning, we introduce a novel Instagram dataset. We make the code and data publicly available.

(4) With quantitative evaluation and user studies via AMT, we demonstrate the effectiveness of three novel features of CSMN and its performance superiority over state-of-the-art captioning models, including [27, 28, 31].

## 2. Related work

**Image Captioning**. In recent years, much work has been published on image captioning, including [3, 4, 9, 12, 20, 22, 28, 31, 33], to name a few. Many proposed captioning models exploit RNN-based decoders to generate a sequence of words from encoded representation of input images. For example, long-term recurrent convolutional networks [3] are one of earliest model to use RNNs for modeling the relations between sequential inputs and outputs. You *et al.* [33] exploit semantic attention to combine top-down and bottom-up strategies to extract richer information from images, and couples it with an LSTM decoder. Compared to such recent progress of image captioning research, it is novel to replace an RNN-based decoder with a sequence memory. Moreover, no previous work has tackled the personalization issue, which is the key objective of this work. We also introduce post completion and hashtag prediction as solid and practical applications of image captioning.

| Dataset | # posts | # users | # posts/user | # words/post |
|---|---|---|---|---|
| caption | 721,176 | 4,820 | 149.6 (118) | 8.55 (8) |
| hashtag | 518,116 | 3,633 | 142.6 (107) | 7.45 (7) |

Table 1. Statistics of *Instagram* dataset. We also show average and median (in parentheses) values. The total unique posts and users in our dataset are $(1,124,815/6,315)$.

**Personalization in Vision and Language Research**. There have been many studies about personalization in computer vision and natural language processing [2, 8, 1, 32, 10, 23]. Especially, Denton *et al*. [2] develop a CNN model that predicts hashtags from image content and user information. However, this work does not formulate the hashtag prediction as image captioning, and not address the post completion. In computer vision, Yao *et al*. [32] propose a domain adaptation approach to classify user-specific human gestures. Almaev *et al*. [1] adopt a transfer learning framework to detect person-specific facial action unit detection. In NLP, Mirkin *et al*. [18] enhance machine translation performance by exploiting personal traits. Polozov *et al*. [23] generate personalized mathematical word problem for a given tutor/student specification by logic programming. Compared to these papers, our problem setup is novel in that personalization issues in image captioning have not been discussed yet.

**Neural Networks with Memory**. Various memory network models have been proposed to enable neural networks to store variables and data over long timescales. Neural Turing Machines [5] use external memory to solve algorithmic problems such as sorting and copying. Later, this architecture is extended to Differential Neural Computer (DNC) [6] to solve more complicated algorithmic problems such as finding shortest path and graph traversal. Weston *et al*. [29] propose one of the earliest memory network models for natural language question answering (QA), and later Sukhbaatar *et al*. [24] modify the network to be trainable in an end-to-end manner. Kumar *et al*. [13] and Miller*et al*. [17] address language QA tasks proposing novel memory networks such as dynamic networks with episodic memory in [13] and key-value memory networks in [17]. Compared to previous memory networks, our CSMN has three novel features as discussed in section 1.

## 3. Dataset

We introduce our newly collected *Instagram* dataset, whose key statistics are outlined in Table 1. We make separate datasets for post completion and hashtag prediction.

### 3.1. Collection of Instagram Posts

We collect image posts from Instagram, which is one of the fastest growing photo-sharing social networks. As a post crawler, we use the built-in hashtag search function provided by Instagram APIs. We select 270 search keywords, which consist of the 10 most common hashtags for each of 27 general categories of Pinterest (*e.g*. design, food, style). We use the Pinterest categories because they are well-defined topics to obtain image posts of diverse users. We totally collect 3,455,021 raw posts from 17,813 users.

Next we process a series of filtering. We first apply language filtering to include only English posts; we exclude the posts where more than 20% of words are not in English based on the dictionary *en.us dict* of *PyEnchant*. We then remove the posts that embed hyperlinks in the body text because they are likely to be advertisement. Finally, if users have more than $\max(15, 0.15 \times \#user\_posts)$ non-English or advertisement posts, we remove all of their posts.

Next we apply filtering rules for the lengths of captions and hashtags. We limit maximum number of posts per user to 1,000, not to make the dataset biased to a small number of dominant users. We also limit minimum number of posts per user to 50, to be sufficiently large to discover users' writing patterns from posts. We also filter out the posts if their lengths are too short or too long. We set 15 as maximum post length because we observe that lengthy posts tend to include irrelevant stuff to the associated pictures. We set 3 as minimum post length because too short posts are likely to include only an exclamation (*e.g. great!*) or a short reply (*e.g. thanks to everyone!*). We use the same rule for hashtag dataset. We observe that lengthy lists of hashtags more than 15 are often too much redundant (*e.g.* #fashionable, #fashionblog, #fashionista, #fashionistas, #fashionlover, #fashionlovers). Finally, we obtain about 721,176 posts for captions and 518,116 posts for hashtags.

### 3.2. Preprocessing

We separately build a vocabulary dictionary $\mathcal{V}$ for each of the two tasks, by choosing the most frequent $V$ words in our dataset. For instance, the dictionary for hashtag prediction includes only most frequent hashtags as vocabularies. We set $V$ to 40K for post completion and 60K for hash prediction after thorough tests. Before building the dictionary, we first remove any urls, unicodes except emojis, and special characters. We then lowercase words and change user names to a @username token.

## 4. The Context Sequence Memory Network

Figure 2 illustrates the proposed *context sequence memory network* (CSMN) model. The input is a query image $I_q$ of a specific user, and the output is a sequence of words: $\{y_t\} = y_1, \ldots, y_T$, each of which is a symbol coming from the dictionary $\mathcal{V}$. That is, $\{y_t\}$ corresponds to a list of hashtags in hashtag prediction, and a post sentence in post generation. Optional input is the context information to be added to memory, such as active vocabularies of a given user. Since both tasks can be formulated as word sequence prediction for a given image, we exploit the same CSMN model while changing only the dictionary. Especially, we too regard hashtag prediction as sequence prediction instead of prediction of a bag of orderless tag words. Since hashtags in a post tend to have strong co-occurrence relations, it is better to take previous hashtags into account to predict a next one. It will be validated by our experimental results.

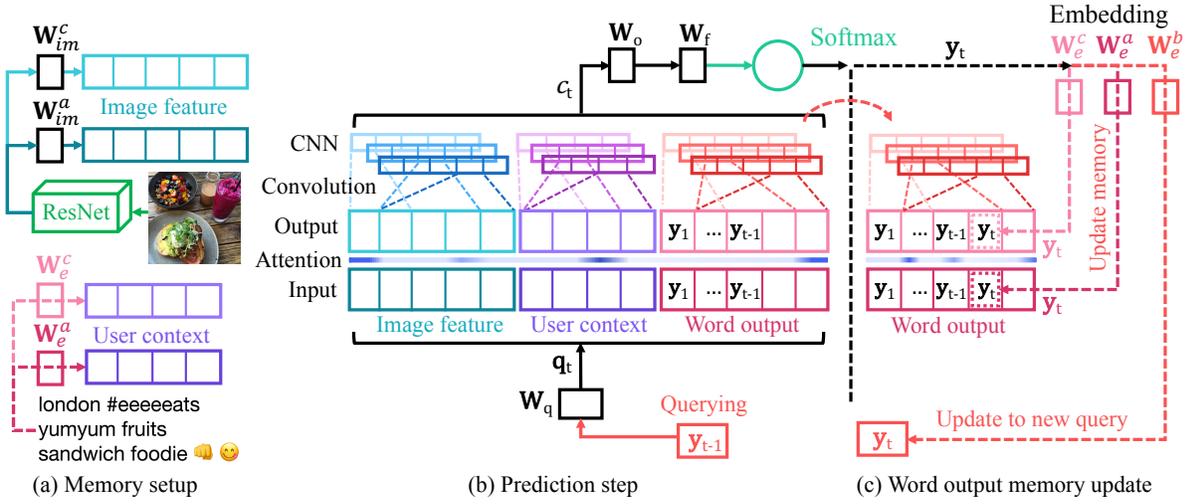

Figure 2. Illustration of the proposed *context sequence memory network* (CSMN) model. (a) Context memory setup using image descriptions and $D$ frequent words from the query user's previous posts (section 4.1). (b) Word prediction at every step $t$ based on the memory state (section 4.2). (c) Update of word output memory once a new output word is generated.

### 4.1. Construction of Context Memory

As in Figure 2(a), we construct the memory to store three types of context information: (i) *image memory* for representation of a query image, (ii) *user context memory* for TF-IDF weighted $D$ frequent words from the query user's previous posts, and (iii) *word output memory* for previously generated words. Following [29], each input to the memory is embedded into input and output memory representation, for which we use superscript $a$ and $c$, respectively.

**Image Memory**. We represent images using ResNet-101 [7] pretrained on the ImageNet 2012 dataset. We test two different descriptions: $(7 \times 7)$ feature maps of res5c layer, and pool5 feature vectors. The res5c feature map denoted by $\mathbf{I}^{r5c} \in \mathbb{R}^{2,048 \times 7 \times 7}$ is useful if a model exploits spatial attention; otherwise, the pool5 feature $\mathbf{I}^{p5} \in \mathbb{R}^{2,048}$ is used as a feature vector of the image. Hence, the pool5 is inserted into a single memory cell, while the res5c feature map occupies 49 cells, on which the memory attention later can focus on different regions of an $(7 \times 7)$ image grid. We will compare these two descriptors in the experiments.

The image memory vector $\mathbf{m}_{im} \in \mathbb{R}^{1,024}$ for the res5c feature is represented by

$$\mathbf{m}^a_{im,j} = \text{ReLU}(\mathbf{W}^a_{im} \mathbf{I}^{r5c}_j + \mathbf{b}^a_{im}), \quad (1)$$
$$\mathbf{m}^c_{im,j} = \text{ReLU}(\mathbf{W}^c_{im} \mathbf{I}^{r5c}_j + \mathbf{b}^c_{im}), \quad (2)$$

for $j = 1, \ldots, 49$. The parameters to learn include $\mathbf{W}^{a,c}_{im} \in \mathbb{R}^{1,024 \times 2,048}$ and $\mathbf{b}^{a,c}_{im} \in \mathbb{R}^{1,024}$. The ReLU indicates an element-wise ReLU activation [19]. For the pool5, we use

$$\mathbf{m}^{a/c}_{im,j} = \text{ReLU}(\mathbf{W}^{a/c}_{im} \mathbf{I}^{p5}_j + \mathbf{b}^{a/c}_{im}). \quad (3)$$

for $j = 1$. In Eq.(3), we simply present two equations for input and output memory as a single one using superscript $a/c$. Without loss of generality, we below derive the formulation assuming that we use the res5c feature.

**User Context Memory**. In a personalized setting where the author of a query image is identifiable, we define $\{u_i\}_{i=1}^{D}$ by selecting $D$ most frequent words from the user's previous posts. We input $\{u_i\}_{i=1}^{D}$ into the user context memory in a decreasing order of scores, in order to exploit CNN later effectively. This context memory improves the model's performance by focusing more on the user's writing style of active vocabularies or hashtags. To build $\{u_i\}_{i=1}^{D}$, we compute TF-IDF scores and select top-$D$ words for a given user. Using TF-IDF scores means that we do not include too general terms that many users commonly use, because they are not helpful for personalization. Finally, the user context memory vector $\mathbf{m}^{a/c}_{us} \in \mathbb{R}^{1,024}$ becomes

$$\mathbf{u}^a_j = \mathbf{W}^a_e \mathbf{u}_j, \mathbf{u}^c_j = \mathbf{W}^c_e \mathbf{u}_j; \mathbf{y}_j; \quad j \in 1, \ldots, D \quad (4)$$
$$\mathbf{m}^{a/c}_{us,j} = \text{ReLU}(\mathbf{W}_h [\mathbf{u}^{a/c}_j] + \mathbf{b}_h), \quad (5)$$

where $\mathbf{u}_j$ is a one-hot vector for $j$-th active word. Parameters include $\mathbf{W}^{a/c}_e \in \mathbb{R}^{512 \times V}$ and $\mathbf{W}_h \in \mathbb{R}^{1,024 \times 512}$. We use the same $\mathbf{W}_h$ for both input and output memory, while we learn separate word embedding matrices $\mathbf{W}^{a/c}_e$.

**Word Output Memory**. As shown in Figure 2(c), we insert a series of previously generated words $y_1, \ldots, y_{t-1}$ into the word output memory, which is represented as

$$\mathbf{o}^a_j = \mathbf{W}^a_e \mathbf{y}_j, \mathbf{o}^c_j = \mathbf{W}^c_e \mathbf{y}_j; \quad j \in 1, \ldots, t-1 \quad (6)$$
$$\mathbf{m}^{a/c}_{ot,j} = \text{ReLU}(\mathbf{W}_h [\mathbf{o}^{a/c}_j] + \mathbf{b}_h). \quad (7)$$

where $\mathbf{y}_j$ is a one-hot vector for $j$-th previous word. We use the same word embeddings $\mathbf{W}^{a/c}_e$ and parameters $\mathbf{W}_h, \mathbf{b}_h$ with user context memory in Eq.(4). We update $\mathbf{m}^{a/c}_{ot,j}$ for every iteration whenever a new word is generated.

Finally, we concatenate the input and memory presentation of all memory types: $\mathbf{M}_t^{a/c} = [\mathbf{m}_{im,1}^{a/c} \oplus \cdots \oplus \mathbf{m}_{im,49}^{a/c} \oplus \mathbf{m}_{us,1}^{a/c} \oplus \cdots \oplus \mathbf{m}_{us,D}^{a/c} \oplus \mathbf{m}_{ot,1}^{a/c} \oplus \cdots \oplus \mathbf{m}_{ot,t-1}^{a/c}]$. We use $m$ to denote the memory size, which is the sum of sizes of three memory types: $m = m_{im} + m_{us} + m_{ot}$.

### 4.2. State-Based Sequence Generation

RNNs and their variants have been widely used for sequence generation via recurrent connections throughout time. However, our approach does not involve any RNN module, but sequentially store all of previously generated words into the memory. It enables to predict each output word by selectively attending on the combinations of all previous words, image regions, and user context.

We now discuss how to predict a word $y_t$ at time step $t$ based on the memory state (see Figure 2(b)). Letting one-hot vector of previous word to $\mathbf{y}_{t-1}$, we first generate an input vector $\mathbf{q}_t$ at time $t$ to our memory network as

$$\mathbf{q}_t = \text{ReLU}(\mathbf{W}_q \mathbf{x}_t + \mathbf{b}_q), \text{ where } \mathbf{x}_t = \mathbf{W}_e^b \mathbf{y}_{t-1}. \quad (8)$$

where $\mathbf{W}_e^b \in \mathbb{R}^{512 \times V}$ and $\mathbf{W}_q \in \mathbb{R}^{1,024 \times 512}$ are learned. Next $\mathbf{q}_t$ is fed into the attention model of context memory:

$$\mathbf{p}_t = \text{softmax}(\mathbf{M}_t^a \mathbf{q}_t), \; \mathbf{M}o_t(*, i) = \mathbf{p}_t \circ \mathbf{M}_t^c(*, i). \quad (9)$$

We compute how well the input vector $\mathbf{q}_t$ matches with each cell of memory $\mathbf{M}_t^a$ by a matrix multiplication followed by a softmax. That is, $\mathbf{p}_t \in \mathbb{R}^m$ indicates the compatibility of $\mathbf{q}_t$ over $m$ memory cells. Another interpretation is that $\mathbf{p}_t$ indicates which part of input memory is important for input $\mathbf{q}_t$ at current time step (*i.e.* to which part of memory the *attention* turns at time $t$ [31]). Next we rescale each column of the output memory presentation $\mathbf{M}_t^c \in \mathbb{R}^{m \times 1,024}$ by element-wise multiplication (denoted by ∘) with $\mathbf{p}_t \in \mathbb{R}^m$. As a result, we obtain the attended output memory representation $\mathbf{M}o_t$, which are decomposed into three memory types as $\mathbf{M}o_t = [\mathbf{m}_{im,1:49}^o \oplus \mathbf{m}_{us,1:D}^{a/c} \oplus \mathbf{m}_{ot,1:t-1}^{a/c}]$.

**Memory CNNs**. We then apply a CNN to the attended output of memory $\mathbf{M}o_t$. As will be shown in our experiments, using a CNN significantly boosts the captioning performance. It is mainly due to that the CNN allows us to obtain a set of powerful representations by fusing multiple heterogeneous cells with different filters.

We define a set of three filters whose depth is 300 by changing window sizes $h = [3, 4, 5]$. We separately apply a single convolutional layer and max-pooling layer to each memory type. For $h = [3, 4, 5]$,

$$\mathbf{c}_{im,t}^h = \text{maxpool}(\text{ReLU}(\mathbf{w}_{im}^h * \mathbf{m}_{im,1:49}^o + \mathbf{b}_{im}^h)) \quad (10)$$

where $*$ indicates the convolutional operation. Parameters include biases $\mathbf{b}_{im}^h \in \mathbb{R}^{49 \times 300}$ and filters $\mathbf{w}_{im}^h \in \mathbb{R}^{[3,4,5] \times 1,024 \times 300}$. Via max-pooling, each $\mathbf{c}_{im,t}^h$ is reduced from $(300 \times [47, 46, 45])$ to $(300 \times [1, 1, 1])$. Finally, we obtain $\mathbf{c}_{im,t}$ by concatenating $\mathbf{c}_{im,t}^h$ from $h = 3$ to 5. We repeat the convolution and maxpooling operation of Eq.(13) to the other memory types as well. As a result, we obtain $\mathbf{c}_t = [\mathbf{c}_{im,t} \oplus \mathbf{c}_{us,t} \oplus \mathbf{c}_{ot,t}]$, whose dimension is $2,700 = 3 \times 3 \times 300$.

Next we compute the output word probability $\mathbf{s}_t \in \mathbb{R}^V$:

$$\mathbf{h}_t = \text{ReLU}(\mathbf{W}_o \mathbf{c}_t + \mathbf{b}_o), \quad (11)$$
$$\mathbf{s}_t = \text{softmax}(\mathbf{W}_f \mathbf{h}_t). \quad (12)$$

We obtain the hidden state $\mathbf{h}_t$ by Eq.(11) with a weight matrix $\mathbf{W}_o \in \mathbb{R}^{2,700 \times 2,700}$ and a bias $\mathbf{b}_o \in \mathbb{R}^{2,700}$. We then compute the output probability $\mathbf{s}_t$ over vocabularies $\mathcal{V}$ by a softmax layer in Eq.(12).

Finally, we select the word that attains the highest probability $y_t = \text{argmax}_{\mathbf{s} \in \mathcal{V}}(\mathbf{s}_t)$. Unless the output word $y_t$ is the EOS token, we repeat generating a next word by feeding $y_t$ into the word output memory in Eq.(6) and the input of Eq.(8) at time step $t + 1$. As a simple post-processing only for hashtag prediction, we remove duplicate output hashtags. In summary, this inference is *greedy* in the sense that the model creates the best sequence by a sequential search for the best word at each time step.

### 4.3. Training

To train our model, we adopt teacher forced learning that provide the correct memory state to predict next words. We use the softmax cross-entropy loss as the cost function for every time step predictions, which minimizes the negative log likelihood from the estimated $y_t$ to its corresponding target word $y_{GT,t}$. We randomly initialize all the parameters with a uniform unit scaling of 1.0 factor: $[\pm \sqrt{3/dim}]$.

We apply mini-batch stochastic gradient descent. We select the Adam optimizer [11] with $\beta_2 = 0.9, \beta_2 = 0.999$ and $\epsilon = 1e - 08$. To speed up training, we use four GPUs for data parallelism, and set a batch size as 200 for each GPU. We earn the best results the initial learning rate is set as 0.001 for all the models. At every 5 epochs, we divide a learning rate by 1.2 to gradually decrease it. We train our models up to 20 epochs.

## 5. Experiments

We compare the performance of our approach with other state-of-the-art models via quantitative measures and Amazon Mechanical Turk (AMT) studies.

### 5.1. Experimental Setting

We use the image of a test post as a query and associated hashtags and text description as groundtruth (GT). For evaluation metrics of hashtag prediction, we compute the F1-score as a balanced average metric between precision and recall between predicted hashtag sets and GT sets: $2(1/\text{precision}+1/\text{recall})^{-1}$. For evaluation measures of post generation, we compute the language similarity between predicted sentences and GTs. We exploit BLEU [21], CIDEr [26], METEOR [14], and ROUGE-r [15] scores. In all measures, higher scores indicate better performance.

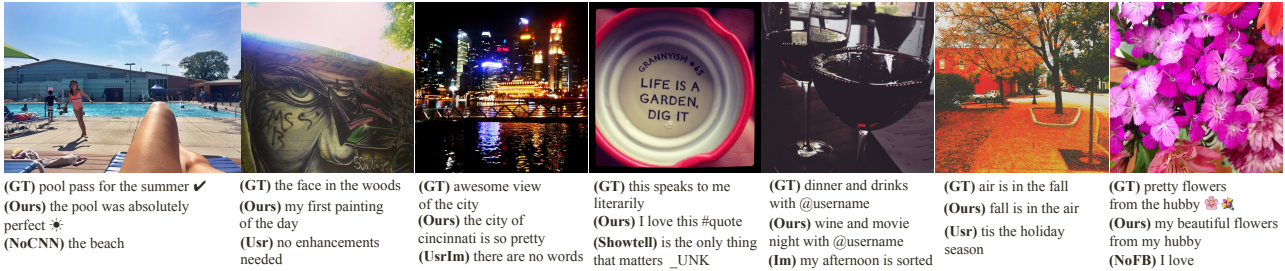

Figure 3. Seven examples of post generation with query images, groundtruths (GT), and generated posts by our method (Ours) and baselines. The @username shows an anonymized user. Most of the predicted texts are relevant and meaningful for the query images.

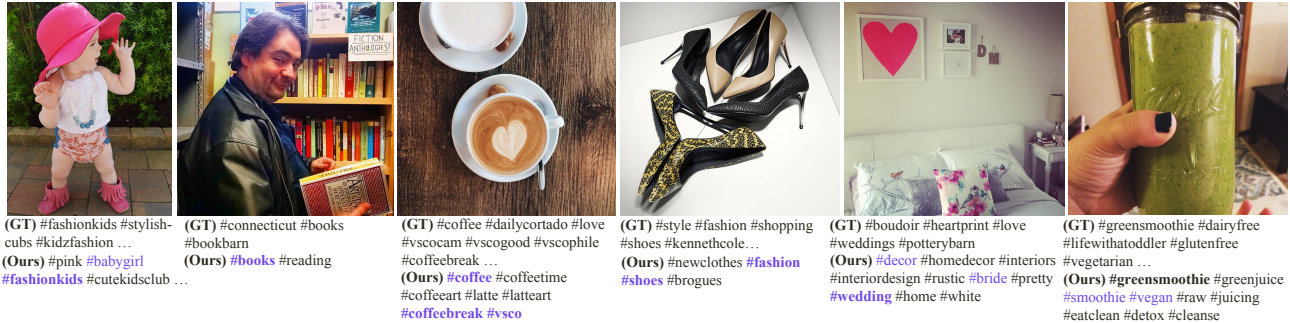

Figure 4. Six examples of hashtag prediction with query images, groundtruths (GT), and our predicted hashtags (Ours). Bold hashtags are correctly matched ones that occur in both (GT) and (Ours). Colored words are the ones that appear in both prediction and context memory.

We randomly split the dataset into 90% for training, 5K posts for test and the rest for validation. We divide the dataset *by users* so that training and test users are disjoint, in order to correctly measure the prediction power of methods. If users's posts exist both in training and test sets, then prediction can be often trivial by simply retrieving their closest posts in the training set.

While some benchmark datasets for image captioning (*e.g.* Flickr30K [34] and MS COCO [16]) have multiple GTs (*e.g.* 5 sentences per image in the MS COCO), our dataset has only one GT post text and hashtag list per test example. Hence, the absolute metric values in this work may be lower than those in these benchmark datasets.

### 5.2. Baselines

As baselines, we select multiple nearest neighbor approaches, one language generation algorithm, two state-of-the-art image captioning methods, and multiple variants of our model. As straightforward baselines, we first test the 1-nearest search by images, denoted by (`1NN-Im`); for a query image, we find its closest training image using the $\ell_2$ distance on ResNet pool5 descriptors, and return its text as prediction. Second, we test the 1-nearest search by users, denoted by (`1NN-Usr`); we find the nearest user whose 60 active vocabularies are overlapped the most with those of the query user, and then randomly select one post of the nearest user. The third nearest neighbor variant denoted by (`1NN-UsrIm`) is to find the 1-nearest image among the nearest user's images, and return its text as prediction.

As a language-only method, we use the sequence-to-sequence model by Vinyals *et al.* [27], denoted by (`seq2seq`). It is a recurrent neural network with three hidden LSTM layers, and originally applied to the language translation. This baseline takes 60 active words of the query user in a decreasing order of TF-IDF weights, and predicts captions. Since this baseline does not use an image for text generation, this comparison quantifies how much the image is important to predict hashtags or text.

We also compare with the two state-of-the-art image captioning methods with no personalization. The first baseline is (`ShowTell`) of [28], which is a multi-modal CNN and LSTM model. The second baseline is the attention-based captioning model of [31] denoted by (`AttendTell`).

We compare different variants of our method (`CSMN-⋆`). To validate the contribution of each component, we exclude one of key components from our model as follows: (i) without the memory CNN in section 4.2 denoted by (`-NoCNN-`), (ii) without user context memory denoted by (`-NoUC-`), and (iii) without feedback of previously generated words to output memory by (`-NoWO-`). That is, the (`-NoCNN-`) quantifies the performance improvement by the use of the memory CNN. The (`-NoUC-`) is the model without personalization; that is, it does not use the information about query users, such as their $D$ active vocabularies. Finally, the (`-NoWO-`) is the model without sequential prediction. For hashtag prediction, the (`-NoWO-`) indicates the performance of separate tag generation instead of sequen-tial prediction of our original proposal. We also test two different image descriptors in section 4.1: (7 × 7) res5c feature maps and pool5 feature vectors, denoted by (-R5C) and (-P5) respectively. Finally, we also evaluate the effects on the sizes of user context memory: (-W20-) and (-W80-) or (-W100-).

| Methods | B-1 | B-2 | B-3 | B-4 | METEOR | CIDEr | ROUGE-L |
|---|---|---|---|---|---|---|---|
| (seq2seq) [27] | 0.050 | 0.012 | 0.003 | 0.000 | 0.024 | 0.034 | 0.065 |
| (ShowTell)* [28] | 0.055 | 0.019 | 0.007 | 0.003 | 0.038 | 0.004 | 0.081 |
| (AttendTell)* [31] | 0.106 | 0.015 | 0.000 | 0.000 | 0.026 | 0.049 | 0.140 |
| (1NN-Im)* | 0.071 | 0.020 | 0.007 | 0.004 | 0.032 | 0.059 | 0.069 |
| (1NN-Usr) | 0.063 | 0.014 | 0.002 | 0.000 | 0.028 | 0.025 | 0.059 |
| (1NN-UsrIm) | 0.106 | 0.032 | 0.011 | 0.005 | 0.046 | 0.084 | 0.104 |
| (CSMN-NoCNN-P5) | 0.086 | 0.037 | 0.015 | 0.000 | 0.037 | 0.103 | 0.122 |
| (CSMN-NoUC-P5)* | 0.079 | 0.032 | 0.015 | 0.008 | 0.037 | 0.133 | 0.120 |
| (CSMN-NoWO-P5) | 0.090 | 0.040 | 0.016 | 0.006 | 0.037 | 0.119 | 0.116 |
| (CSMN-R5C) | 0.097 | 0.034 | 0.013 | 0.006 | 0.040 | 0.107 | 0.110 |
| (CSMN-P5) | **0.171** | **0.068** | **0.029** | **0.013** | **0.064** | **0.214** | **0.177** |
| (CSMN-W20-P5) | 0.116 | 0.041 | 0.018 | 0.007 | 0.044 | 0.119 | 0.123 |
| (CSMN-W100-P5) | 0.109 | 0.037 | 0.015 | 0.007 | 0.042 | 0.109 | 0.112 |

Table 2. Evaluation of post generation between different methods for the Instagram dataset. As performance measures, we use language similarity metrics (BLEU, CIDEr, METEOR, ROUGE-L). The methods with [*] use no personalization.

| Methods | F1 score | |
|---|---|---|
| (seq2seq) [27] | 0.132 | 0.085 |
| (ShowTell)* [28] | 0.028 | 0.011 |
| (AttendTell)* [31] | 0.020 | 0.014 |
| (1NN-Im)* | 0.049 | 0.110 |
| (1NN-Usr) | 0.054 | 0.173 |
| (1NN-UsrIm) | 0.109 | 0.380 |
| (CSMN-NoCNN-P5) | 0.135 | 0.310 |
| (CSMN-NoUC-P5)* | 0.111 | 0.076 |
| (CSMN-NoWO-P5) | 0.117 | 0.244 |
| (CSMN-R5C) | 0.192 | 0.340 |
| (CSMN-P5) | **0.230** | **0.390** |
| (CSMN-W20-P5) | 0.147 | 0.349 |
| (CSMN-W80-P5) | 0.135 | 0.341 |

Table 3. Evaluation of hashtag prediction. We show test results for *split by users* in the left and *split by posts* in the right.

### 5.3. Quantitative Results

Table 2 and 3 summarize the quantitative results of post generation and hashtag prediction, respectively. Since algorithms show similar patterns in both tasks, we below analyze the experimental results together.

First of all, according to most metrics in both tasks, our approach (CSMN-*) significantly outperforms baselines. We can divide the algorithms into two groups with or without personalization; the latter includes (ShowTell), (AttendTell), (1NN-Im), and (CSMN-NoUC-P5), while the former comprises the other methods. Our (CSMN-NoUC-P5) ranks the first among the methods with no personalization, while the (CSMN-P5) achieves the best overall. Interestingly, using a pool5 feature vector as image description that occupies only a single memory slot leads better performance than using (7 × 7) res5c feature maps with 49 slots. It is mainly due to that attention learning quickly becomes harder with a larger dimension of image representation. Another reason could be that users do not tent to discuss in the level of details about individual (7 × 7) image grids, and thus a holistic view of the image content is sufficient for prediction of users' text.

We summarize other interesting observations as follows. First, among the baselines, the simple nearest neighbor approach (1NN-UsrIm) turns out to be the strongest candidate. Second, our approach becomes significantly worsen, if we remove one of key components, such as memory CNN, personalization, and sequential prediction. Third, among the tested memory sizes of user context, the best performance is obtained with 60. With larger memory sizes, attention learning becomes harder. Moreover, we choose the size of 60 based on the statistics of our dataset; with a too large size, there are many empty slots, which also make attention learning difficult. Finally, we observe that post generation is more challenging than hashtag prediction. It is due to that the expression space of post generation is much larger because the post text includes any combinations of words, emojis, and symbols.

Given that Instagram provides a function of hashtag automation from previous posts when writing a new post, we test another dataset split for hashtag prediction. That is, we divide dataset *by posts* so that each user's posts are included in both training and test sets. We call this as *split by posts*, while the original split as *split by users*. We observe that, due to the automation function, many posts in our training and test set have almost identical hashtags. This setting is highly favorable for (1NN-UsrIm), which returns the text of the closest training image of the query user. Table 3 shows the results of *split by users* in the left and *split by posts* in the right. Interestingly, our (CSMN-P5) works better than (1NN-UsrIm) even in the setting of *split by posts*, although its performance margin (*i.e.* 0.01 in F1 score) is not as significant as in the *split by users* (*i.e.* 0.121).

### 5.4. User Studies via Amazon Mechanical Turk

We perform AMT tests to observe general users' preferences between different algorithms for the two post automation tasks. For each task, we randomly sample 100 test examples. At test, we show a query image and three randomly sampled complete posts of the query user as a personalization clue, and two text descriptions generated by our method and one baseline in a random order. We ask turkers to choose more relevant one among the two. We obtain answers from three different turkers for each query. We select the variant (CSMN-P5) as a representative of our method, because of its best quantitative performance. We compare with three baselines by selecting the best method in each group of 1NNs, image captioning, and language-only methods: (1NN-UsrIm), (ShowTell), and (seq2seq).

Table 4 summarize the results of AMT tests, which validate that human annotators significantly prefer our results to those of baselines. Among the baselines, (1NN-UsrIm) is preferred the most, given that the performance gap with our approach is the smallest. These results coincide with those of quantitative evaluation in Table 2 and 3.

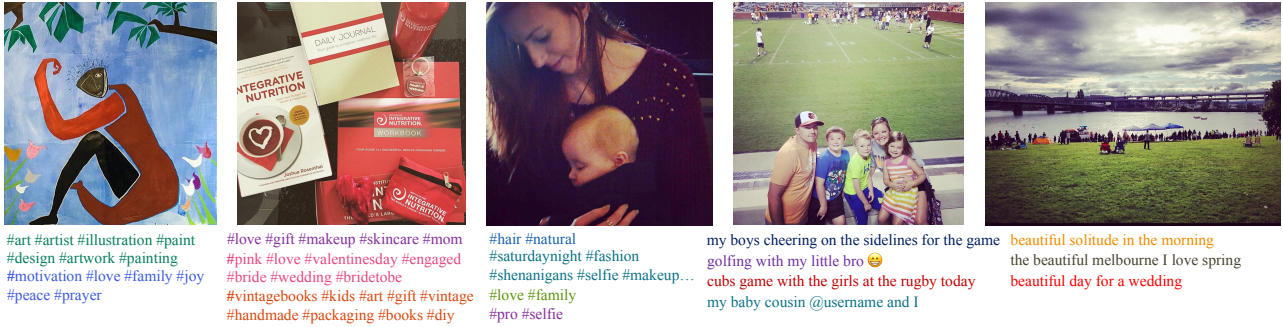

Figure 5. Three examples of hashtag prediction and two examples of post prediction with query images and multiple predictions by different users (shown in different colors). Predicted results vary according to query users, but still are relevant and meaningful for the query images.

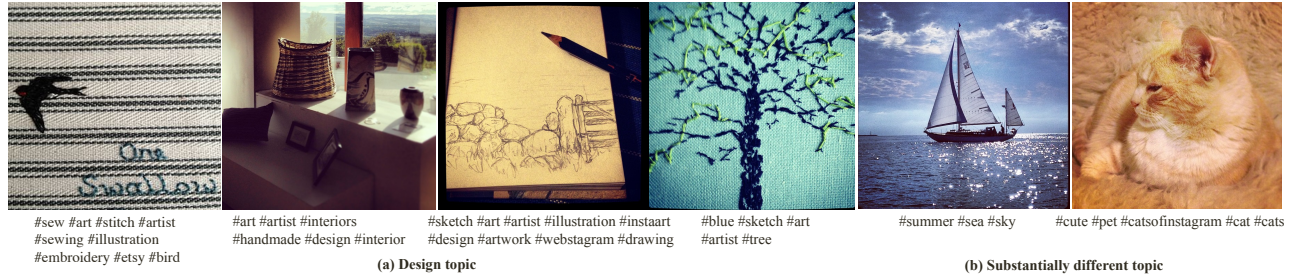

(a) Design topic    (b) Substantially different topic

Figure 6. Six examples of hashtag prediction for a single user whose most of posts are about *design*. (a) For design-related query images, our CSMN predicts relevant hashtags for the design topic. (b) For the query images of substantially different topics, our CSMN is also resilient to predict meaningful hashtags.

| Hashtag Prediction | | | |
|---|---|---|---|
| vs. Baselines | (1NN-UsrIm) | (Showtell) | (seq2seq) |
| (CSMN-P5) | **67.0** (201/300) | **88.0** (264/300) | **81.3** (244/300) |
| Post Generation | | | |
| (CSMN-P5) | **73.0** (219/300) | **78.0** (234/300) | **81.3** (244/300) |

Table 4. AMT preference results for the two tasks between our methods and three baselines. We show the percentages of responses that turkers vote for our approach over baselines.

### 5.5. Qualitative Results

Figure 3 illustrates selected examples of post generation. In each set, we show a query image, GT, and generated text description by our method and baselines. In many of Instagram examples, GT comments are hard to correctly predict, because they are extremely diverse, subjective, and private conversation over a variety of topics. Nonetheless, most of predicted text descriptions are relevant to the query images. Moreover, our CSMN model is able to appropriately use normal words, emojis, and even mentions to other users (anonymized by @username). Figure 4 shows examples of hashtag prediction. We observe that our hashtag prediction is robust even with a variety of topics, including profiles, food, fashion, and interior design.

Figure 5 shows examples of how much hashtag and post prediction vary according to different users for the same query images. Although predicted results change in terms of used words, they are relevant and meaningful for the query images. Figure 6 illustrates the variation of text predictions according to query images for the same user. We first select a user whose most of posts are about *design*, and then obtain prediction by changing the query images. For design-related query images, the CSMN predicts relevant hashtags for the design topic (Figure 6(a)). For the query images of substantially different topics, our CSMN is also resilient to predict relevant hashtags (Figure 6(b)).

### 6. Conclusions

We proposed the context sequence memory networks (CSMN) as a first personalized image captioning approach. We addressed two post automation tasks: hashtag prediction and post generation. With quantitative evaluation and AMT user studies on nearly collected Instagram dataset, we showed that our CSMN approach outperformed other state-of-the-art captioning models. There are several promising future directions that go beyond this work. First, we can extend the CSMN model for another interesting related task such as *post commenting* that generates a thread of replies for a given post. Second, since we dealt with only Instagram posts in this work, we can explore data in other social networks such as Flickr, Pinterest, or Tumblr, which have different post types, metadata, and text and hashtag usages.

**Acknowledgements**. This research is partially supported by Hancom and Basic Science Research Program through National Research Foundation of Korea (2015R1C1A1A02036562). Gunhee Kim is the corresponding author.

## A. Why using memory CNN can be helpful?

While conventional memory networks cannot model the structural ordering unless time embedding is added to the model, we propose to exploit the memory CNN to model the structural ordering with much stronger representation power. More specifically, for the word output memory, the CNN is useful to represent the sequential order of generated words. For the user context memory, the CNN can correctly capture the importance order of the context words, given that we feed the user's frequent words into the context memory in a decreasing order of TF-IDF weighted scores (rather than putting them in a random order).

For example, suppose that a user can have active words related to *fashion*, *street* and *landscape* in the user context memory. If the *fashion*-related words are at the top of the user context memory, and the *street* is located between *fashion*-related words, this user can be modeled to be interested in *street fashion*. On the other hand, if the *landscape*-related words are at the top of the memory, and the *street* is located between *landscape*-related words, this user can be modeled to be interested in *landscape* since the meaning of *street* can be interpreted similarly as *landscape*. Without the memory CNN, it can be difficult to distinguish between these two different uses of *street*.

## B. Details of Dataset Collection

**27 general categories of Pinterest.** For dataset collection, we select 270 search keywords by gathering the 10 most common hashtags for each of the following 27 general categories of Pinterest: celebrities, design, education, food, drink, gardening, hair, health, fitness, history, humor, decor, outdoor, illustration, quotes, product, sports, technology, travel, wedding, tour, car, football, animal, pet, fashion and worldcup.

**Dataset statistics.** We show the cumulative distribution functions (CDFs) of the words for captions and hashtags in Figure 7. Based on the statistics, we set the vocabulary size for captions as 40K and for hashtags as 60K. For captions, top 40K most frequent words take 97.3% of all the word occurrences, indicating sufficient coverage for the vocabulary usage in the Instagram. On the other hand, hashtags in Instagram show extremely diverse neologism. The dictionary of 60K vocabularies cover only 84.31%, but we set 60K because the increase of the CDF is very slow even with increasing the dictionary size.

We also show some statistics of the active vocabularies per user for captions and hashtags in Table 5. We set the sizes of user context memory based on these values. If the memory size is too small, then the memory cannot correctly capture the users' writing style. On the other hand, if the memory size is too large, the learning becomes hard and performance decreases. We set the sizes so that they roughly correspond to the sum of mean and standard deviation values).

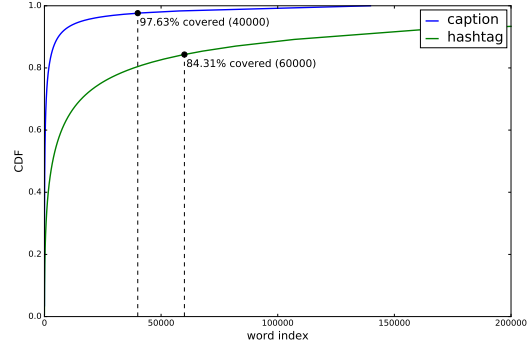

Figure 7. Cumulative distribution functions (CDFs) of the words for captions and hashtags in the *Instagram* dataset. For captions (hashtags), the top 40K (60K) most frequent words take 97.63% (84.31%) of all the word occurrences of the dataset.

| Dataset | # words/user (Mean) | # words/user (Stddev) |
|---------|---------------------|------------------------|
| caption | 22 | 34 |
| hashtag | 31 | 34 |

Table 5. Statistics of active vocabularies per user (mean and standard deviation) in *Instagram* dataset.

## C. Model Details

In the main draft, we describe our model assuming that we use the res5c features $\mathbf{I}^{r5c} \in \mathbb{R}^{2,048 \times 7 \times 7}$ for image representation. Here we discuss some formulation changes when we use the pool5 features $\mathbf{I}^{p5} \in \mathbb{R}^{2,048}$.

For memory CNNs, we represent its output of memory for res5c features as follows (*i.e.* Eq.(10) of the main draft).

$$\mathbf{c}^h_{im,t} = \text{maxpool}(\text{ReLU}(\mathbf{w}^h_{im} * \mathbf{m}^o_{im,1:49} + \mathbf{b}^h_{im})) \quad (13)$$

When we use pool5 features, Eq.(13) is changed to

$$\mathbf{c}^h_{im,t} = \mathbf{w}^h_{im} \mathbf{m}^o_{im,1} + \mathbf{b}^h_{im}. \quad (14)$$

where $\mathbf{b}^h_{im} \in \mathbb{R}^{1,800}$, $\mathbf{w}_{im}h \in \mathbb{R}^{1,024 \times 1,800}$, and $\mathbf{c}^h_{im,t} \in \mathbb{R}^{1,800}$. After experiments, we found out that adding ReLu on Eq.(14) slightly improves performance.

Finally, the memory output concatenation for res5c features, which is represented by

$$\mathbf{c}_t = [\mathbf{c}_{im,t} \oplus \mathbf{c}_{us,t} \oplus \mathbf{c}_{ot,t}], \quad (15)$$

is also changed to, when we use pool5 features,

$$\mathbf{c}_t = \mathbf{c}_{im,t} + [\mathbf{c}_{us,t} \oplus \mathbf{c}_{ot,t}]. \quad (16)$$

The dimension of $\mathbf{c}_t$ is changed from $2,700 = 3 \times 3 \times 300$ for rec5c features to $1,800 = 2 \times 3 \times 300$ for pool5 features.

## D. Training Details

Although our model can take variable-length sequences as input, to speed up training, it is better for a minibatch to consist of sentences with the same length. Therefore, we randomly group training samples to a set of minibatches, each of which have the same length as possible. We then randomly shuffle a batch order so that short and long minibatches are mixed. It empirically leads a better training as a curriculum learning in [35] proposes.